\documentclass{article} 
\usepackage{iclr2025_conference,times}

\usepackage{amsmath,amsfonts,bm}

\def\eqref#1{equation~\ref{#1}}

\def\1{\bm{1}}

\DeclareMathAlphabet{\mathsfit}{\encodingdefault}{\sfdefault}{m}{sl}
\SetMathAlphabet{\mathsfit}{bold}{\encodingdefault}{\sfdefault}{bx}{n}

\usepackage{hyperref}
\usepackage{url}

\usepackage{multirow}
\usepackage{graphicx}
\usepackage{subcaption}
\usepackage{enumitem}
\usepackage{amsmath}
\usepackage{amssymb}
\setlist[itemize]{leftmargin=5mm}
\usepackage{amsmath, amssymb, mathtools, nccmath}
\DeclarePairedDelimiter{\nint}\lfloor\rceil

\newcommand{\Hquad}{\hspace{0.5em}} 
\usepackage{adjustbox}
\usepackage{float}
\usepackage{tabularray}
\usepackage{tabularx}
\usepackage{sidecap}
\usepackage{comment}
\usepackage{wrapfig}
\usepackage{booktabs,tabularx, colortbl} 
\usepackage{lipsum}

\let\svthefootnote\thefootnote
\newcommand\freefootnote[1]{%
  \let\thefootnote\relax%
  \footnotetext{#1}%
  \let\thefootnote\svthefootnote%
}

\newcommand{\kv}{KV-Cache}
\newcommand{\lr}{low-rank}

\newcommand{\palu}{\textit{Palu}}

\title{Palu: KV-Cache Compression with Low-Rank Projection}

\author{Chi-Chih Chang$^{1,3 \ast \dagger}$\quad  
Wei-Cheng Lin$^{1 \ast}$  \quad 
Chien-Yu Lin$^{2 \ast}$  \quad \\
\textbf{Chong-Yan Chen$^{1}$}\quad  
\textbf{Yu-Fang Hu$^{1}$} \quad 
\textbf{Pei-Shuo Wang$^{1}$}\quad
\textbf{Ning-Chi Huang$^{1}$}\quad \\
\textbf{Luis Ceze$^{2}$}\quad 
\textbf{Mohamed S. Abdelfattah$^{3}$}\quad
\textbf{Kai-Chiang Wu$^{1}$} 
\\$^1$National Yang Ming Chiao Tung University\quad
$^2$University of Washington\quad
$^3$Cornell University \\
}

\iclrfinalcopy 
\begin{document}

\maketitle
\def\thefootnote{$\ast$}\footnotetext{Equal contribution}
\def\thefootnote{$\dagger$}\footnotetext{Correspondence to: Chi-Chih Chang,  cc2869@cornell.edu}
\begin{abstract}

Post-training \kv{} compression methods typically either sample a subset of effectual tokens or quantize the data into lower numerical bit width. 
However, these methods cannot exploit redundancy in the hidden dimension of the KV tensors. 
This paper presents a hidden dimension compression approach called \palu{}, a \kv{} compression framework that utilizes low-rank projection to reduce inference-time LLM memory usage. 
Palu decomposes the linear layers into low-rank matrices, caches compressed intermediate states, and reconstructs the full keys and values on the fly. 
To improve accuracy, compression rate, and efficiency, \palu{} further encompasses (1) a medium-grained low-rank decomposition scheme, (2) an efficient rank search algorithm, (3) low-rank-aware quantization compatibility enhancements, and (4) optimized GPU kernels with operators fusion. 
Extensive experiments with popular LLMs show that \palu{} compresses \kv{} by 50\%, while maintaining strong accuracy and delivering up to \textbf{1.89$\times$ speedup} on the RoPE-based attention module. 
When combined with quantization, \palu{}'s inherent quantization-friendly design yields small to negligible extra accuracy degradation, while saving additional memory than quantization-only methods and achieving up to \textbf{2.91$\times$ speedup} for the RoPE-based attention.
Moreover, it maintains comparable or even \textbf{better accuracy (up to 1.19 lower perplexity)} compared to quantization-only methods. 
These results demonstrate \palu{}'s superior capability to effectively address the efficiency and memory challenges of LLM inference posed by \kv{}. Our code is publicly available at: \href{https://github.com/shadowpa0327/Palu}{https://github.com/shadowpa0327/Palu}

\end{abstract}
\section{Introduction}
\label{sec:intro}

Large language models (LLMs) have propelled AI into new applications and capabilities, providing a high-level intelligence that previous machine learning (ML) models could not achieve.
To speed up inference, caching the Key-Value states (\kv{}) in memory is a simple yet effective technique. 
However, the size of the \kv{} can grow rapidly, straining memory capacity and bandwidth especially with long context lengths \citep{LongContextChallenge}; further, the memory-bounded nature of the decoding stage limits inference speed when loading \kv{} data \citep{MemoryWall}. 
Therefore, \kv{} compression has become a central research topic for running LLMs efficiently.

Although emerging attention mechanisms such as Multi-Query Attention (MQA) \citep{mqa}, Group-Query Attention (GQA) \citep{ainslie2023gqa} and Multi-head Latent Attention (MLA) \citep{deepseekai2024deepseekv2} can reduce KV-Cache size, it either requires model pre-training or has a significant impact on model's accuracy when converting from traditional Multi-Head Attention (MHA) \citep{chen2024optimisedgroupedqueryattentionmechanism}.
In contrast, post-training \kv{} compression techniques offer an alternative approach to advance efficiency for existing models.
Among various \kv{} compression methods, quantization \citep{kivi, kvquant} and token eviction \citep{h2o, attention_sinks} stand out as effective strategies to reduce the memory footprint of \kv{}. 

Quantization methods aim to reduce the bit-width used to represent each piece of data, while token eviction techniques focus on retaining a partial set of \kv{}. 
However, both methods neglect the hidden dimensions of the KV-Cache, where substantial redundancy often resides. 
To capitalize on this untapped potential, we introduce \palu{}, a post-training \kv{} compression framework that leverages low-rank projection to reduce the hidden dimension of KV tensors, offering an additional and orthogonal compression dimension to existing quantization and token eviction methods.

\begin{wrapfigure}[18]{hr}{0.51\textwidth}
\begin{center}
    \vspace{-5pt}
    \includegraphics[width=0.50\textwidth, trim = 0 2.0cm 0 2.7cm]{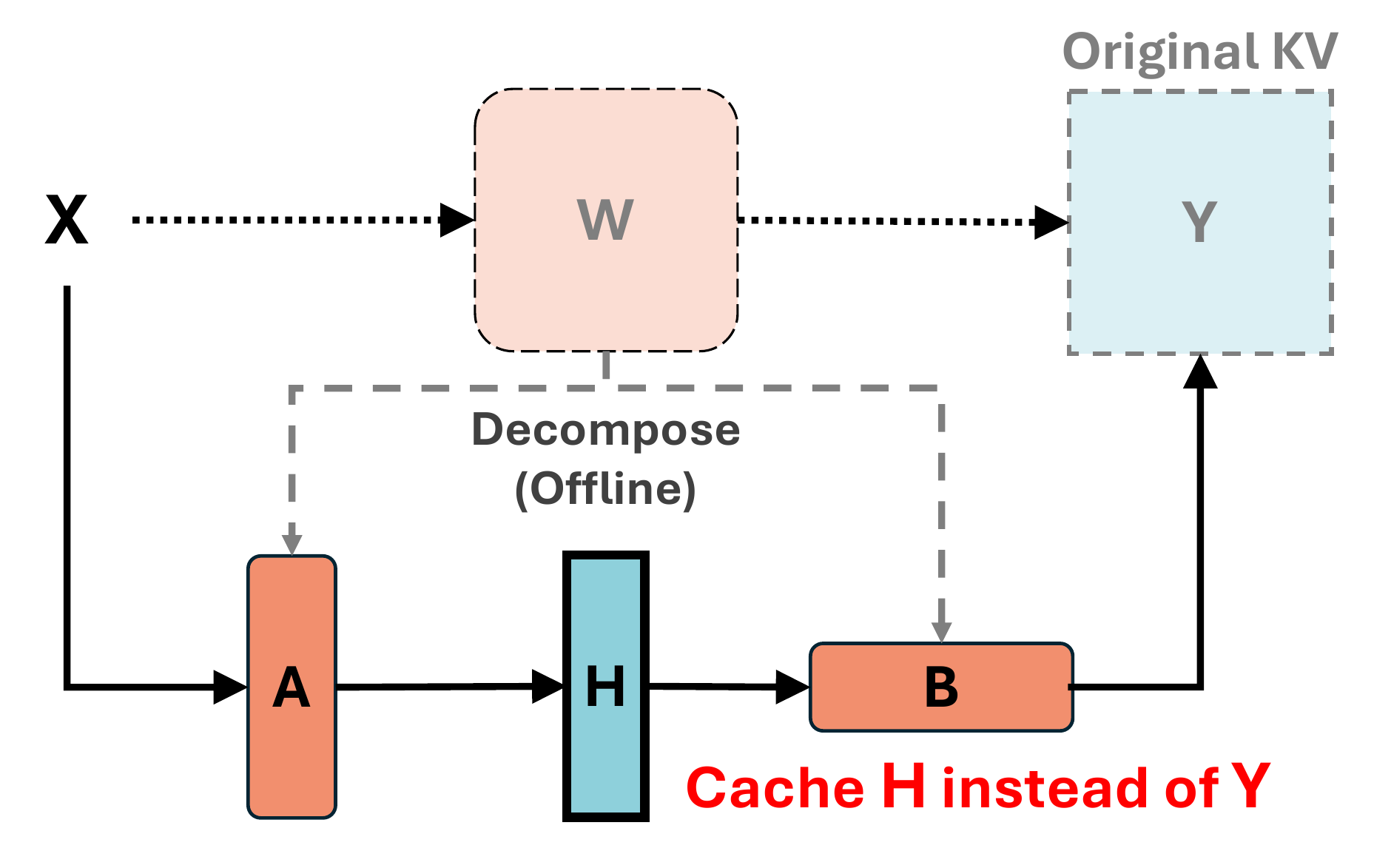}
\end{center}
\caption{
        \palu{}’s \textit{low-rank projection method} for \kv{} reduction. A weight matrix \textbf{W} of linear projection is decomposed into two low-rank matrices. Input \textbf{X} is down-projected to latent representation \textbf{H}, which is cached. \textbf{Y} can be reconstructed from \textbf{H} using the up-projection matrix \textbf{B}.
    }
\label{fig: concept overview}
\end{wrapfigure}

A naive way to utilize \lr{} projection for compressing the \kv{} is by directly mapping cached matrices into \lr{} space \citep{PCA2, Galore}. 
However, this approach imposes an unacceptably heavy overhead of computing the decomposition matrices during runtime. 
To avoid this, \palu{} \textit{statically decomposes the Key and Value-projection weight matrices} and \textit{caches the latent representations} of the \lr{} decomposition (see Fig.~\ref{fig: concept overview}).
This innovative design enables \palu{} to reduce memory while mitigating the runtime overhead of \kv{} low-rank decomposition.

In designing an effective decomposition strategy for attention modules with multiple attention heads, we observed a clear trade-off between accuracy and reconstruction overhead. 
Decomposing the projection matrices across all attention heads together improves accuracy by preserving global information, but this approach significantly increases reconstruction costs. 
On the other hand, decomposing each head separately reduces reconstruction overhead but leads to a higher loss in accuracy. 
To address this, \palu{} introduces a medium-grained, group-head low-rank decomposition that strikes a balance between accuracy and reconstruction efficiency.

For LLMs, each linear projection module has a different sensitivity to compression \citep{LASER, asvd}. To exploit the sensitivity and improve accuracy, we design an efficient \textit{rank search algorithm} based on Fisher information \citep{Fisher, GroupedFisher}. Our algorithm automatically assigns a higher rank for important matrices and lower ranks for less critical ones, boosting accuracy at the same overall \kv{} compression rate. 

In addition to its low-rank decomposition, \palu{} is compatible with quantization techniques. We found that low-rank decomposition can introduce severe outliers in the latent representation, which significantly hinders accurate low-bit quantization. Although the Hadamard transformation has been shown to be effective for outlier elimination in recent studies \citep{quip, quarot, spinquant, Quamba}, its integration often introduces computational overhead during runtime. However, \palu{}'s inherent matrix pair structure makes it highly compatible with this technique, allowing the transformation matrices to be seamlessly fused into the forward and backward matrices, effectively mitigating outliers without impacting runtime efficiency. 

We evaluate \palu{} on widely used LLMs and benchmarks. Our experiments demonstrate that \palu{} maintains strong zero-shot accuracy and perplexity with up to 50\% \lr{} compression. Moreover, when combining \lr{} compression with quantization, \palu{} achieves an impressive \textbf{over 91.25\% compression (11.4$\times$ reduction)} and yields a \textit{significantly lower perplexity of 1.19} than KVQuant \citep{kvquant}, a state-of-the-art \kv{} quantization method, which only achieves an 87.5\% compression rate.

For latency evaluation, under a 50\% \kv{} compression rate without quantization, \palu{} demonstrates up to \textit{1.89$\times$ and 2.2$\times$ speedup} for RoPE-based and non-RoPE attention modules.
When integrated with quantization, \palu{} achieves up to \textit{2.91$\times$ and 6.17$\times$ acceleration} on RoPE-based and non-RoPE attention, respectively.
These results underscore \palu{}’s ability to significantly reduce \kv{} memory footprint while boosting inference efficiency for LLMs.

\newpage
Our key contributions include: 
\begin{itemize}
\vspace{-5pt}
\setlength{\itemsep}{-0.2em}
    \item \textit{\palu{}}, a new post-training \kv{} compression framework that caches \textit{low-rank latent representations} of Key and Value states.
    \item \textit{Group-head low-rank decomposition (G-LRD)}, an optimization for balancing accuracy and reconstruction efficiency.
    \item An \textit{automated rank search} \textit{algorithm} for adaptively assigning ranks to each decomposed matrix, given a target compression rate.
    \item A co-designed \textit{quantization compatibility optimization} that eliminates \lr{}-induced outliers and imposes zero runtime overhead.
\end{itemize}

\vspace{-4pt}
\section{Background}

\subsection{Multi-Head Attention Mechanism}
\label{Prelimary: MHA}
The multi-head attention (MHA) mechanism \citep{attentionIsAllyouNeed} is a core component of the transformer architecture. Given a new input token $\mathbf{x} \in \mathbb{R}^{d}$, an MHA with $n$ heads projects the input into multiple queries, keys, and values using weight matrices $\mathbf{W}^q_{i}$, $\mathbf{W}_i^k$, and $\mathbf{W}_i^v$, respectively, for each head \(i\), as shown by 

\begin{equation}
    \mathbf{q}_i = \mathbf{x} \mathbf{W}_i^q, \Hquad \mathbf{k}_{i} = \mathbf{x} \mathbf{W}^{k}_i, \Hquad  \mathbf{v}_{i} = \mathbf{x} \mathbf{W}^{v}_i.
\end{equation}

Here, \(\mathbf{k}_{i}\) and \(\mathbf{v}_{i}\) represent the key and value at time step \(t\) for head \(i\). We can then compute the attention score for each head \(i\) and the corresponding attention output as 
\begin{equation}
\label{eq:Attention Score}
\begin{aligned}
\mathbf{p}_{t,i} = \text{Softmax} \left( \frac{{\mathbf{q}_i} \mathbf{K}_{i}^T}{\sqrt{d_h}} \right), \Hquad  \mathbf{a}_{i}= \mathbf{p}_{i} \mathbf{V}_{i},
\end{aligned}
\end{equation}
where \(\mathbf{K}_{i}\) and \(\mathbf{V}_{i}\) denote the concatenation of current and all previous keys and values corresponding to the $i$-th head.
The final MHA output is obtained by concatenating the outputs of all heads and then applying the out-projection layer $\mathbf{W}_o$, as shown by  

\begin{equation}
\text{MHA}(\mathbf{x}) = \sum_{i=1}^{h} \mathbf{a}_{i} \mathbf{W}^o_i = \sum_{i=1}^{h} (\mathbf{p}_{i} \mathbf{V}_{i}) \mathbf{W}^o_i,
\end{equation}
where \(\mathbf{W}^o_i \in \mathbb{R}^{d_h \times d}\) represents the submatrices of the out-projection matrix for each head $i$.


\subsection{Singular Value Decomposition (SVD)}
\label{subsec:SVD}
SVD \citep{PCA2}  is a commonly used technique for computing the \lr{} approximation for a given matrix.
We now introduce our specific use of SVD for \palu{}'s default \lr{} decomposition method.

Given a weight matrix $\mathbf{W} \in \mathbb{R}^{m \times n}$, SVD decomposes $\mathbf{W}$ into three matrices: $\mathbf{W} = \mathbf{U} \boldsymbol{\Sigma} \mathbf{V}^T$. Here, $\mathbf{U}$ and $\mathbf{V}$ are orthogonal matrices containing the left and right singular vectors, respectively. The matrix $\boldsymbol{\Sigma}$ is a diagonal matrix that consists of singular values.
We describe the decomposition as 


\begin{equation*}
\label{eq:low-rank}
    \mathbf{W} \approx \mathbf{A}\mathbf{B}, \Hquad   \mathbf{A} = \mathbf{U}_r \sqrt{\boldsymbol{\Sigma}_r}, \Hquad  \mathbf{B} = \sqrt{\boldsymbol{\Sigma}_r} \mathbf{V}_r^T, 
\end{equation*}
where  $\mathbf{A} \in \mathbb{R}^{m \times r}$, $\mathbf{B} \in \mathbb{R}^{r \times n}$,
$\boldsymbol{\Sigma}_r \in \mathbb{R}^{r \times r}$. $\boldsymbol{\Sigma}_r$ is a diagonal matrix containing the largest $r$ singular values, and $\mathbf{U}_r$, $\mathbf{V}_r^T$ are corresponding singular vectors truncated from $\mathbf{U}$ and $\mathbf{V}^T$. This truncation and subsequent matrix formation let us approximate weight matrix $\mathbf{W}$ with two low-rank matrices $\mathbf{A}$ and $\mathbf{B}$, thereby reducing the storage by $\frac{mr + rn}{mn}$.

\begin{figure*}[t]
\centering
    \includegraphics[width=0.9\textwidth]{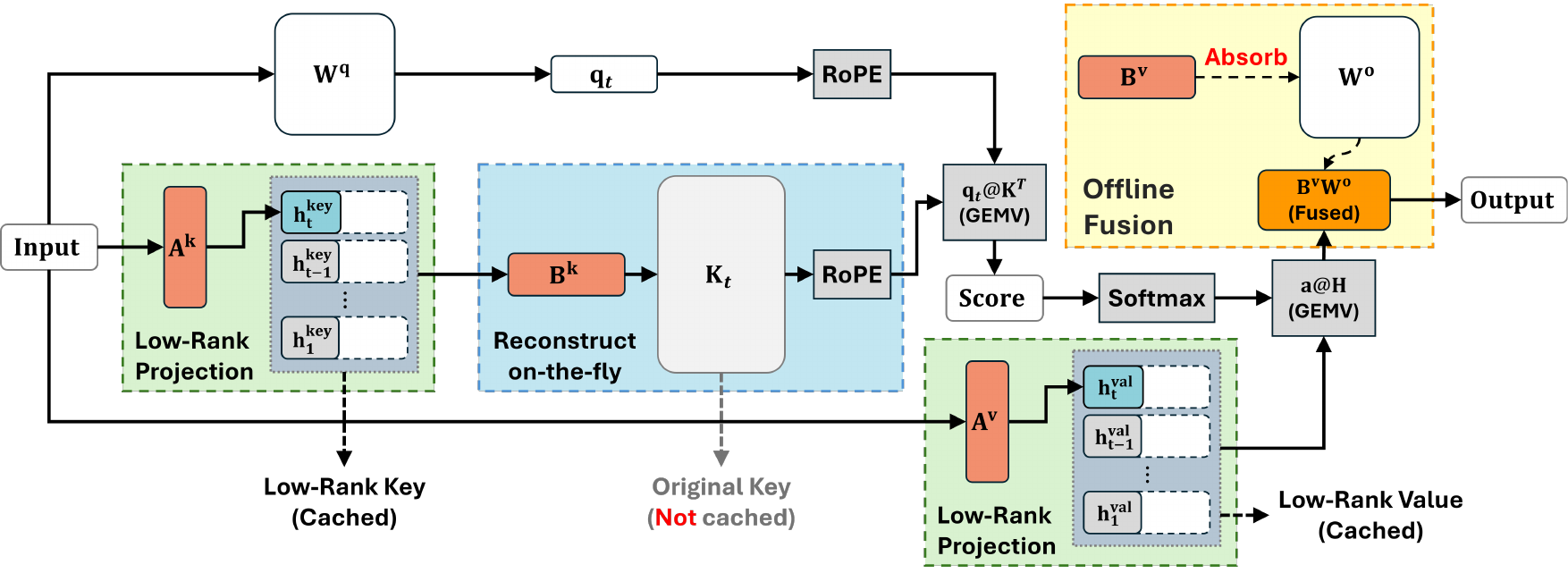}
    \caption{\palu{} uses low-rank decomposition $(\mathbf{W} \approx \mathbf{A}\mathbf{B})$ to project the key (or value) to a lower-dimensional latent representation $(\mathbf{h})$, thereby reducing the size of the \kv{}. 
    The original key $(\mathbf{K}_t)$ is reconstructed on-the-fly with $\mathbf{B}^k$, and $\mathbf{B}^v$ is fused into $\mathbf{W}^o$ to avoid reconstruction overhead. The fusion also reduces the computational burden for output projection.}
    \label{fig: framework}
\end{figure*}

\vspace{-5pt}
\section{The Palu Framework}

\subsection{Compressing the \kv{} via Low-Rank Projection}
To apply low-rank projection more efficiently than directly decomposing the \kv{} during runtime, \palu{} uses SVD to decompose the Key and Value projection matrices. 
This approach is based on the observation that low-rank decomposition rewrites the linear projection layer from $\mathbf{y}=\mathbf{x}\mathbf{W}$ into 
$\mathbf{y}=\mathbf{x}\mathbf{A} \mathbf{B}$.
Here, $\mathbf{A} \in \mathbb{R}^{d \times r}$ is the low-rank projection matrix, and $\mathbf{B} \in \mathbb{R}^{r \times d}$ is the reconstruction matrix derived by SVD. 
The forward process first \textit{down-projects} the input token $\mathbf{x}\in\mathbb{R}^{d}$ into a low-dimensional latent space $\mathbf{h}\in\mathbb{R}^{r}$ and then \textit{up-projects} it back to the original space:
\begin{equation}
\mathbf{h} = \mathbf{A}\mathbf{x}, \quad \mathbf{y} = \mathbf{B}\mathbf{h}
\end{equation}
This two-step process lets \palu{} (1) store the lower dimension latent representation instead of the origin key and value states, and (2) reconstruct them during decoding.

\subsubsection{Integration with the Attention Mechanism and Offline Matrix Fusion} 
\label{sec:Attention Integration and fusion}
We now describe how \palu{} decomposes the key and value linear layers for the attention mechanism.
For each attention head $i$, \palu{} applies SVD and maps the key-projection matrix $\mathbf{W}^{k}_i$ and value-projection matrix $\mathbf{W}^{v}_i$ into $\mathbf{A}^{k}_i \mathbf{B}^{k}_i$ and $\mathbf{A}^{v}_i \mathbf{B}^{v}_i$. 

Based on the formula of attention output in Eq. \ref{eq:Attention Score}, \palu{} absorbs the reconstruction matrix $\mathbf{B}^{v}_i$ into the output projection matrix $\mathbf{W}_i^o$ offline:

\begin{equation}
\begin{aligned}
\label{eq:fused vo}
\mathbf{a}_i\mathbf{W}^o_i &= (\mathbf{p}_i \mathbf{V}_i )\mathbf{W}^o_i = (\mathbf{p}_i \mathbf{H}^{v}_{i} \mathbf{B}^{v}_i) \mathbf{W}^o_i = \mathbf{p}_i \mathbf{H}^{v}_{i} (\mathbf{B}^{v}_i \mathbf{W}^o_i) \\
\end{aligned}
\end{equation}

Such fusion lets \palu{} skip the explicit reconstruction of the full value vectors, reduce the number of matrix multiplications, and improve efficiency. 
A similar approach applies for calculating attention scores. 
Matrix $\mathbf{B}^{k}_i$ can be fused into the query projection matrix $\mathbf{W}_i^q$ offline, as shown by 

\begin{equation}
\begin{aligned}
\label{eq:fused qk}
\mathbf{q}_i \mathbf{K}_i^T &= \mathbf{q}_i(\mathbf{H}^k_i \mathbf{B}^k_i)^T = \mathbf{x}_t \mathbf{W}^q_i (\mathbf{B}^k_i)^T (\mathbf{H}^k_i)^T  = \mathbf{x}_t \Bigl( \mathbf{W}^q_i (\mathbf{B}^k_i)^T \Bigr) (\mathbf{H}^k_i)^T. \\
\end{aligned}
\end{equation}
Here, \(\mathbf{B}^{k}_i \in \mathbb{R}^{r \times d_h}\) and \(\mathbf{W}^q_i \in \mathbb{R}^{d \times d_h}\), so the fused matrix \((\mathbf{W}^q_i (\mathbf{B}^k_i)^T)\) has size \(\mathbb{R}^{d \times r}\). This fusion boosts computational efficiency by reducing the matrix dimension during attention score calculation. 

\subsubsection{Compability with Positional Embedding} 
\label{sec:rope}
Recent LLMs, such as the Llama family, apply Rotary Positional Embedding (\textit{i.e.,} RoPE \citep{RoPE}) onto the Query and Key states prior to their multiplication. 
The non-linear nature of these positional embeddings prevents the matrix fusion of attention scores, as outlined in Eq. \ref{eq:fused qk}.
To address this, \palu{} dynamically reconstructs the keys from latent representations during decoding. 
We enhance reconstruction efficiency with a custom-designed GPU kernel, detailed in Sec. \ref{sec:kernel impl} and further described in Appendix \ref{appendix: kernel impl}.

Note that for some positional embedding methods, such as ALiBi \citep{alibi}, or new attention mechanisms, such as MLA \citep{deepseekai2024deepseekv2}, positional embedding is not directly applied to the Key states.
Consequently, the fusion described in Eq. \ref{eq:fused qk} remains valid. 
For these non-RoPE attention modules, \palu{} achieves greater speedup compared to RoPE-based attention, as their reconstruction can be avoided with matrix fusion.
A detailed workflow of \palu{} in the context of Llama’s RoPE-based attention is illustrated in Fig. \ref{fig: framework}.

\subsection{Decomposition Granularity}

\label{subsec:decompose_granularity}
\subsubsection{Multi-Head Low-Rank Decomposition} 

We name the per-head decomposition scheme in Sec. \ref{sec:Attention Integration and fusion} as \textit{multi-head \lr{} decomposition (M-LRD)}.
We found M-LRD often causes a non-negligible accuracy degradation (discussed further in Sec. \ref{sec:exp_main_results}), possibly because SVD fails to capture the common information shared across heads.
Therefore, alternative approaches are needed to preserve model accuracy.



\subsubsection{Joint-Head Low-Rank Decomposition} 
An alternative approach is to jointly decompose weight matrices for all heads.
By considering the combined weight matrix \(\mathbf{W}_{\text{joint}} = [\mathbf{W}_{1},\mathbf{W}_{2},\dots,\mathbf{W}_{n} ] \in \mathbb{R}^{d \times (d_h \cdot n_h)}\),
we can perform a single low-rank decomposition
$\mathbf{W}_{\text{joint}} \approx \mathbf{A}_{\text{joint}} \mathbf{B}_{\text{joint}}$, 
where \(\mathbf{A}_{\text{joint}} \in \mathbb{R}^{d \times r_\text{joint}}\) and \(\mathbf{B}_{\text{joint}} \in \mathbb{R}^{r_\text{joint} \times (d_h \cdot n_h)}\). 
We call this scheme \textit{joint-head \lr{} decomposition (J-LRD)}.

J-LRD has the advantage of preserving the common principal components shared among different heads. 
This occurs because SVD is particularly effective at capturing the dominant components when applied to a larger, combined matrix, resulting in a more accurate approximation. 


For J-LRD, the joint latent representation 
shared among all heads can be computed with
$\mathbf{h}_{\text{joint}} = \mathbf{x}\mathbf{A}_{\text{joint}}$.
During decoding, the original states for each head can be reconstructed via
\begin{equation*}
\begin{aligned}
\bigl[\mathbf{y}_{1},\dots,\mathbf{y}_{n}\bigr] &= \mathbf{h}_{\text{joint}}\mathbf{B}_{\text{joint}}.
\end{aligned}
\end{equation*}

\paragraph{High Inference Overhead.}  Despite better preserving model accuracy, J-LRD introduces \textit{significant computational and memory overhead} during decoding. Specifically, the total number of floating point operations (FLOPs) to reconstruct the Key or Value state of one head now becomes $r_{\text{joint}}\cdot d_h\cdot n$. Assuming the same size as the total low-rank latent representations (\textit{i.e.,} $r_{\text{joint}} = \sum_{i=1}^{n}{r_i}$), 
the total reconstruction cost is $n$ times higher than M-LRD, whose total FLOPs is $r_i\cdot d_h \cdot n$.
When considering the matrix fusion in Sec. \ref{sec:Attention Integration and fusion}, 
the fused matrix of J-LRD has a size of $r_{joint}\cdot d\cdot n$, which is also $n$ times larger than M-LRD, leading to substantial higher memory consumption.


\begin{figure*}[thp]
    \centering
    \includegraphics[width=0.9\textwidth]{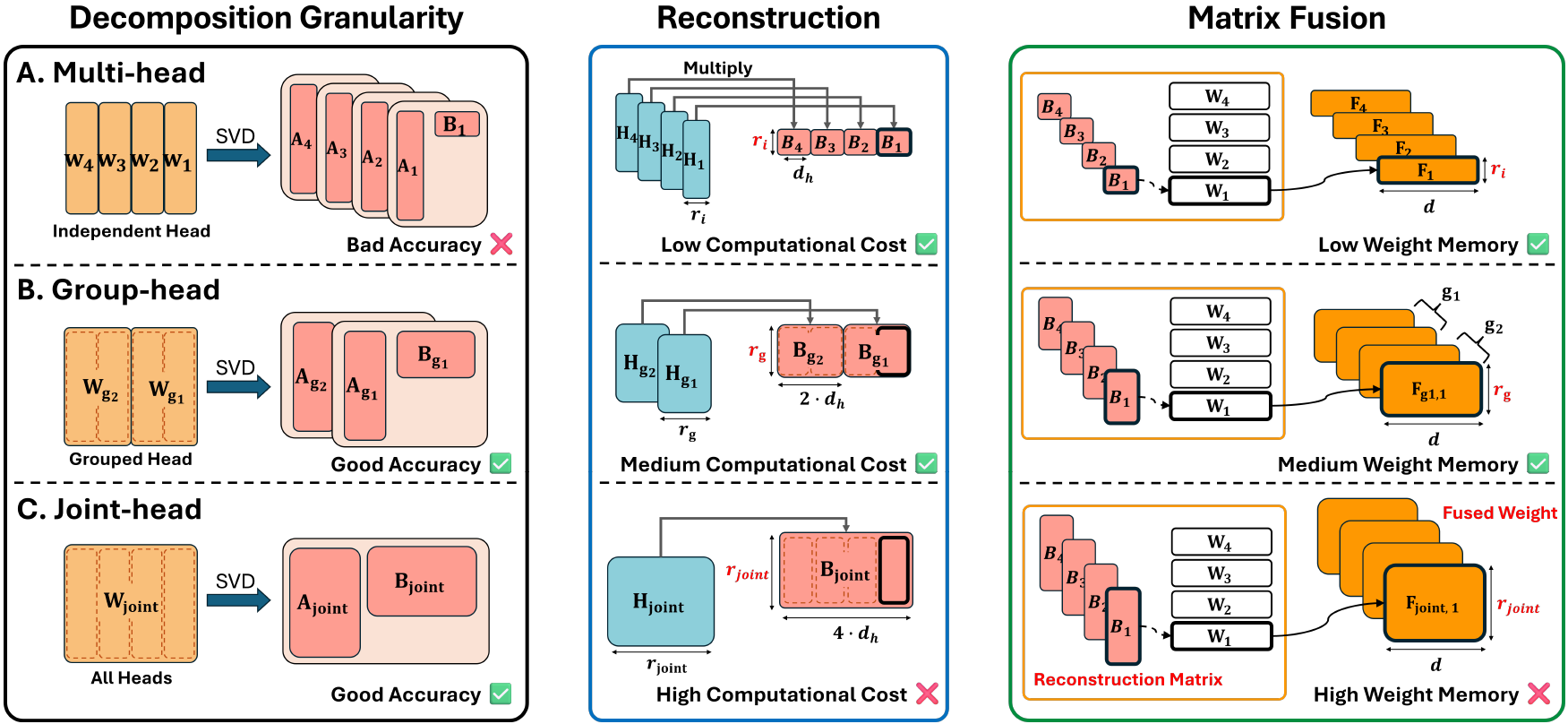}
    \caption{Performing decomposition at different granularities. Jointly decomposing multiple heads can achieve higher accuracy. Assuming the same total size of the latent representations (\textit{i.e.}, $4\cdot r_i=2\cdot r_g=r_{\text{joint}}$), the FLOPs for reconstruction overhead in joint-head decomposition schemes are 4 times larger than those in multi-head ones.}
    \label{fig:group_low_rank}
\end{figure*}

\subsubsection{Group-Head Low-Rank Decomposition} 
To balance the trade-off between accuracy and reconstruction cost, we propose \textit{group-head low-rank decomposition} (G-LRD). 
G-LRD decomposes the matrices for a group of heads together.
With combined weight matrices, it captures shared information within each group while limiting computational overhead and preserving accuracy.

To illustrate the G-LRD process, consider the weight matrices for a group of $s$ heads, 
$\mathbf{W}_{g_j} = \bigl[\mathbf{W}_{j, 1} \dots \mathbf{W}_{j, s}\bigr]$, where $\mathbf{W}_{g_j} \in \mathbb{R}^{d \times (d_h \cdot s)}$.
We \lr{} decompose \mbox{$\mathbf{W}_{g_j} \approx \mathbf{A}_{g_j} \mathbf{B}_{g_j}$}, 
where \(\mathbf{A}_{g_j} \in \mathbb{R}^{d \times r_g}\) and \(\mathbf{B}_{g_j} \in \mathbb{R}^{r_g \times (d_h \cdot s)}\). 
The latent representation shared among attention heads in the same group can be computed as $\mathbf{h}_{\text{g}_j} = \mathbf{x} \mathbf{U}_{\text{g}_j}$.
During decoding, the original key or value for each head can be reconstructed via 
\begin{equation*}
    [\mathbf{y}_{j, 1} \dots \mathbf{y}_{j, s}] = \mathbf{h}_{\text{g}_j} \mathbf{B}_{\text{g}_j}.
\end{equation*}

The FLOPs for reconstructing the keys and values for each head in G-LRD is $r_g \cdot d_h \cdot n_g$, 
where \(n_g=\frac{n}{s}\) is the number of groups. 
Comparing the cost to J-LRD and assuming the same total rank size (\(r_g \cdot n_g = r_\text{joint}\)), G-LRD reduces the reconstruction cost by $n_g$. 
Similarly, G-LRD also reduces the fused matrix size by $n_g$.
To sum up, G-LRD offers a middle ground between computation overhead and approximation accuracy.
We illustrate M-LRD, J-LRD and G-LRD in Fig. \ref{fig:group_low_rank}.
Please refer to Appendix \ref{append:lr_weight_size} for further discussions on the costs of different decomposition granularities.


\subsection{Automatic Rank Allocation}

To allocate an ideal rank size to the decomposition target, it is crucial to accurately estimate the importance of the target matrix (\textit{e.g.,} grouped weights). In \palu{}, we identify \textbf{Fisher information} \citep{Fisher, GroupedFisher} as an accurate approximator since it can quantify the amount of information for each parameter. 
We then employ the sum of Fisher information to estimate the importance of the weight matrix of each linear layer~\citep{abdelfattahzcp}. 

Assuming that the compression sensitivity is proportional to Fisher information, we determine the rank for each weight matrix by computing the ratio of its Fisher information to the total Fisher information across all decomposition targets. 
We use this ratio to allocate the compression rate (\textit{i.e.,} rank level $r$), ensuring that more important layers retain higher rank levels.
For a detailed ablation study on our Automatic Rank Allocation, please refer to Appendix \ref{append:detailed_key_value_rank}.

\subsection{Quantization Compatibility}
\label{sec:Low-Rank Aware Quantization}

\begin{wrapfigure}[10]{r}{0.55\textwidth}
\centering
\vspace{-2.3em}
     \includegraphics[width=.55\textwidth, trim = 0 2.5cm 0 3.0cm]{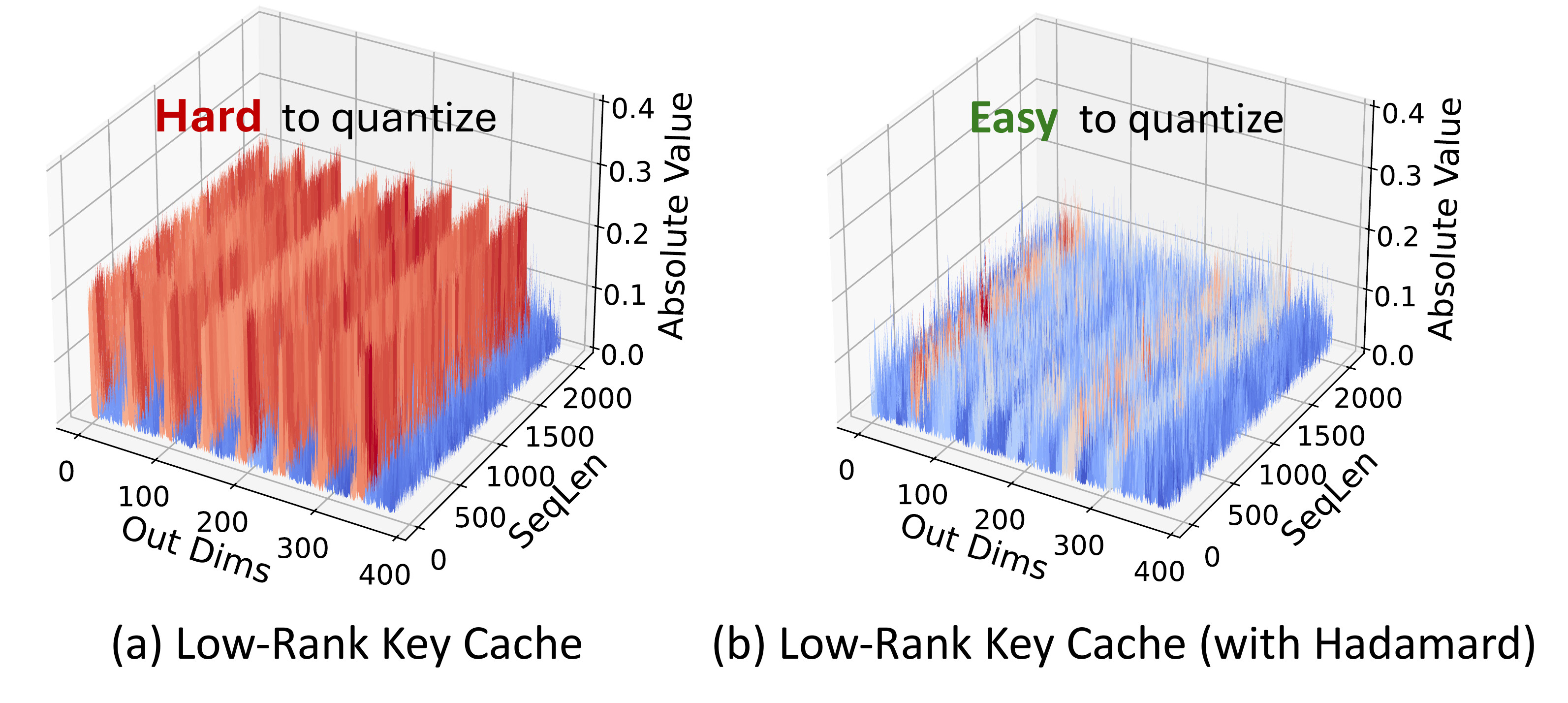}
\vspace{-8pt}
\caption{
        Activation distribution of the low-rank key caches at the $4^{th}$ Llama-2 attention layer.
    }
\label{fig:quantization}
\end{wrapfigure}

We integrate quantization into \palu{} to compress the \kv{} further. We observe that low-rank compressed latent representations have severe outliers, which limit quantization applicability in \palu{}. Unlike natural outliers described in previous \kv{} quantization literature \citep{kivi, kvquant}, these outliers are induced by SVD-based low-rank factorization.

Fig. \ref{fig:quantization} (a) shows the distribution of low-rank compressed key states from a layer of Llama-2 with G-LRD. Repeating outlier patterns appear at the beginning of each decomposed group because SVD arranges larger eigenvalues in the initial rows or columns, resulting in rapidly descending values in the latent representation. This pattern stretches the data distribution and hurts quantization accuracy.

Inspired by recent LLM quantization literature \citep{quarot, quip}, we apply the Walsh-Hadamard transform (WHT, \citeauthor{WHT}) to eliminate outliers (Fig. \ref{fig:quantization} (b)), 
enabling a high quantization accuracy. 
However, this transformation introduces an extra matrix multiplication with associated runtime overhead. 
Unlike earlier methods \citep{quarot} that must apply online WHT when quantizing \kv{}, we optimize this process by integrating the Hadamard matrix into low-rank decomposed weights with no additional compute overhead, as described by

\begin{equation}
    \mathbf{W} \approx \mathbf{AB} = (\mathbf{AR}) (\mathbf{R}^T\mathbf{B}) = \mathbf{\hat{A}\hat{B}},
\end{equation}
where $\mathbf{R}$ is the Hadamard matrix. 
This optimization allows \palu{} to integrate the proposed \lr{} compression technique with low-bit quantization.
Our experiments show that, on top of the \lr{} compression, our quantization method only \textit{negligibly increases perplexity}, even at extreme levels such as \textit{3-bit or 2-bit} with a simple per-token quantization scheme (see Sec. \ref{sec:exp_quantization}).

\section{Experiments}

\subsection{Experiments Setup}

\paragraph{Models and Tasks.} We evaluate \palu{} on four LLM families, Llama-2 \citep{llama2}, Llama-3 \citep{dubey2024llama3herdmodels}, Mistral \citep{mistral} and LongChat \citep{li2023long}.
For accuracy evaluation, we measure perplexity on
the WikiText-2 \citep{wikitext2} and C4 \citep{c4} datasets and use LM-Evaluation-Harness \citep{eval-harness} to measure zero-shot accuracy on six common sense tasks. We also evaluate long context accuracy on 16 tasks in LongBench \citep{LongBench}. Unless specification, we refer to baseline as a model with non-compressed \kv{}. 
See Appendix \ref{append:exp_setting_details} for further details on the dataset and settings.


\paragraph{Compression Settings.} 
We implemented \palu{} based on the Huggingface library \citep{Huggingface}. Decomposition of the Key and Value projection layers was performed using the truncation-aware SVD method proposed by SVD-LLM \citep{svdllm}. Unless otherwise specified, \palu{}'s results are G-LRD with a group size of 4 (gs-4), with equal rank size for each group. To calculate Fisher information in rank searching, we used 2048 random samples from Wikitext-2, each with a sequence length of 1024. 
For quantization integration in \palu{}, we use a simple per-token, asymmetric integer quantization.
For evaluation on quantization results, we compare \palu{} to advanced \kv{} quantization methods, including Atom \citep{atom}, KVQaunt \citep{kvquant}, and KIVI \citep{kivi}.
Refer to Sec. \ref{sec:related work} for a brief summary of these methods.

\paragraph{GPU Kernels Implementation.} 
\label{sec:kernel impl}
To efficiently compute the attention score with RoPE in \palu{}, we implemented a customized kernel in Triton~\citep{tillet2019triton}. Our kernel fuses the key reconstruction, applying RoPE, and the final multiplication with query in a single kernel call, minimizing data movement (See Appendix \ref{appendix: kernel impl}). For quantization integration, we implemented kernels in CUDA for matrix fusion cases on both attention score and attention output (refer to Sec. \ref{sec:Attention Integration and fusion} and Fig. \ref{fig: framework}). Our kernel fuses the dequantization process and the follow-up multiplication with low-rank compressed keys or values, enabling efficient processing on quantized latent \kv{}. When evaluating speedup with quantization, we compare to the non-compressed baseline and KIVI \citep{kivi}, which we use their official code\footnote{https://github.com/jy-yuan/KIVI} in our experiments.

\subsection{Results with Different Decomposition Granularity}
\label{sec:exp_main_results}

We evaluate perplexity and zero-shot accuracy of \palu{} with a \textbf{50\% low-rank compression rate} using M-LRD, G-LRD, and J-LRD on Llama-2-7B and Llama-3-8B-Instruct, and present the results in Table \ref{tab:ppl_and_zero_shot}.

\begin{table}[hb]
\centering
\caption{Perplexity and zero-shot accuracy of \palu{} at 50\% compression rate.
}
\resizebox{35em}{!}{
\begin{tabular}{c | l | rr | ccccccc}
\toprule
\multirow{2}{*}{Model}      & \multicolumn{1}{c|}{\multirow{2}{*}{Method}} & \multicolumn{2}{r|}{Perplexity $\downarrow$}        & \multicolumn{7}{c}{Zero-Shot Accuracy (\%) $\uparrow$}     \\
                            & \multicolumn{1}{c|}{}                        & \multicolumn{1}{r}{Wiki2} & \multicolumn{1}{r|}{C4} & OBQA  & Hella & PIQA  & ARC-e & ARC-c & Wino  & Avg.  \\ \midrule
\multirow{4}{*}{Llama-2-7B}  & Baseline                                     & 5.47                      & 7.26                    & 44.20 & 76.00 & 78.07 & 76.30 & 46.42 & 69.30 & 65.05 \\ \cmidrule{2-11} 
                            & J-LRD                                        & 5.62                      & 7.75                    & 45.40 & 75.57 & 77.48 & 75.97 & 45.31 & 69.22 & 64.82 \\
                            & G-LRD                                        & 6.01                      & 9.82                    & 43.60 & 73.39 & 76.33 & 73.02 & 42.57 & 66.77 & 62.61 \\
                            & M-LRD                                        & 6.75                      & 12.01                   & 39.60 & 65.35 & 74.76 & 67.17 & 35.24 & 64.64 & 57.79 \\ \midrule
\multirow{4}{*}{Llama-3-8B-Inst} & Baseline                                 & 8.28                      & 13.01                   & 43.20&	75.80&	78.62&	81.61&	56.83&	71.90&	67.99 \\ \cmidrule{2-11} 
                            & J-LRD                                        & 9.12                      & 15.90                   & 43.40&	73.20&	76.50&	79.63&	51.96&	72.45&	66.19 \\
                            & G-LRD                                        & 10.11                     & 17.87                   & 42.60&   70.36&  76.06&  76.30&  48.99&  72.38& 64.45 \\
                            & M-LRD                                        & 12.38                     & 23.02                   & 38.80&	63.04&	73.67&	69.78&	42.58&	62.51&	58.40 \\ \bottomrule
\end{tabular}}
\label{tab:ppl_and_zero_shot}
\end{table}
\paragraph{Perplexity Evaluation.} 
As Table \ref{tab:ppl_and_zero_shot} shows, for the Llama-2-7B model, \palu{}'s M-LRD method fails to maintain a low perplexity at a 50\% compression rate. In contrast, despite having a high recomputation cost, J-LRD significantly outperforms M-LRD and achieves a 5.62 perplexity on WikiText-2.

For G-LRD, which still maintains a low computation cost, yields a 6.01 perplexity on Wikitext-2, showing a great balance between model accuracy and compression overheads. The same trend is observed in the Llama-3-8B model as well. More results Llama-2-13B can be found in ppendix \ref{appendix:more_results}. 

\paragraph{Zero-shot Evaluation Results.} Similar to the perplexity evaluation, the J-LRD method demonstrates the best performance for the zero-shot accuracy on Llama-2-7B, with only a 0.23\% average accuracy degradation. M-LRD method results in the lowest average performance, with a 7.26\% drop in accuracy compared to the baseline. In comparison, G-LRD only has a 2.4\% average accuracy decline, offering a sweet spot between model accuracy and compression overheads again.

\subsection{Results of Quantization Integration}
\label{sec:exp_quantization}

\begin{wraptable}{r}{6cm}
\vspace{-4em}
\caption{Quantization perplexity and \kv{} size for Llama-2-7B on WikiText-2. For perplexity, sequence length is 4096. \kv{} size is demonstrated for 128K sequence length.}
\centering
\vspace{-7pt}
\resizebox{15.5em}{!}{
\begin{tabular}{ l | c | c | c | c }
        \toprule
        Method & Bit & PPL &  \begin{tabular}{@{}c@{}}\kv{} \\ Size (GB) \end{tabular} & \begin{tabular}{@{}c@{}}Comp. \\ Rate \end{tabular} \\
        
        \midrule
        Baseline         & 16 & 5.12  & 64.0 & -      \\
        \cmidrule{1-5} 
        Palu-30\%        & 16 & 5.25  & 44.8 & 30\%   \\ 
        Palu-50\%        & 16 & 5.63  & 32.0 & 50\%   \\
        \cmidrule{1-5}         
        Atom             & 3  & 6.15  & 12.6 & 80.32\% \\
        KVQuant          & 3  & 5.35  & 12.0 & 81.25\% \\
        Palu-30\%        & 3  & \textbf{5.33}  & \textbf{8.4}  & \textbf{86.87\%} \\
        Palu-50\%        & 3  & 5.77  & 6.0  & 90.63\% \\
        \cmidrule{1-5}
        Atom             & 2 & 117.88 & 8.6  & 86.56\% \\
        KVQuant          & 2 &   6.95 & 8.0  & 87.50\% \\
        Palu-30\%        & 2 & \textbf{5.76} & \textbf{5.6} & \textbf{91.25\%} \\
        Palu-50\%        & 2 & \textbf{6.41} & \textbf{4.0} & \textbf{93.75\%} \\
        \bottomrule
\end{tabular}}

\vspace{-9pt}
\label{tab:quantization}
\end{wraptable}

Table \ref{tab:quantization} showcases the impact of quantization on perplexity and \kv{} size when combined with \palu{}.
With 3-bit quantization, \palu{} incurs only a \textbf{slight 0.08 and 0.23 perplexity increase} at 30\% and 50\% low-rank compression rate.
These demonstrate a minimal accuracy trade-off for significant compression gains compared to the 16-bit baseline.

Notably, at 2-bit quantization, \palu{} decisively outperforms the state-of-the-art KVQuant method, \textbf{reducing perplexity by 1.19 and 0.54}, while further slashing memory usage by 30\% and 50\%. These results establish \palu{} with quantization as a superior \kv{} compression method.

\subsection{Evaluation on Long Context Datasets}

To access \palu{}'s ability for long-context scenarios, we evaluate baseline, KIVI and \palu{}'s accuracy on LongBench \citep{LongBench}.
Here we evaluate on Mistral-7B and LongChat-7B models, which have up to 32K context length. We report the average score for each task type separately, as well as the overall average across all 16 tasks. The results are shown in Table \ref{tab:longbench_appendix}.

As Table \ref{tab:longbench_appendix} indicates, we find that at a 50\% low-rank compression level, \palu{} is relatively difficult to fully preserve accuracy. However, at a 30\% compression level, \palu{} achieves only a minor average accuracy drop ($<1\%$) compared to the baseline for both models. 
 
Furthermore, \palu{} can quantize the low-rank latent \kv{} down to 3 bits, with less than 1\% further accuracy degradation. 
Overall, \palu{} maintains a strong 40.77\% and 34.33\% average accuracy for Mistral-7B and LongChat-7B, with an impressive 7.59x compression ratio.

When compared to KIVI \citep{kivi}, \palu{} achieves a similar accuracy, while having an additional 30\% compression rate from low-rank.

Notably, \palu{} does not require the complex grouped quantization and mixed-precision techniques employed by KIVI, resulting a high inference efficiency (see Sec. \ref{exp: latency} for details).

\vspace{-5pt}
\subsection{Latency Evaluation}
\vspace{-2pt}
\label{exp: latency}
In this section, we provide latency and speedup evaluation, using Llama-2-7B as the base model. We measure latency on a single RTX 4090 GPU and compare \palu{} to the FP16 and KIVI-4-bit baselines.

We evaluate \palu{}'s latency at a 50\% compression rate, where we set compression rates for key and value to 75\% and 25\%, respectively. 
This allocation is based on our observations from the rank allocation results (see Appendix \ref{append:detailed_key_value_rank} for details).

For FP16 baseline, we use the default implementation from HuggingFace.
For KIVI, we use the CUDA kernels from its official repository.
We do not apply FlashAttention \citep{flashattention} for an apple-to-apple comparison to our baselines.
Due to the small memory capacity of RTX 4090 GPU, we adopt a 4-bit quantization \citep{frantar2024marlinmixedprecisionautoregressiveparallel} for the weights of all linear layers.
Our results are the average of 100 runs.

\begin{table*}[tbhp!]
    \centering
    \caption{Experiment results on Longbench. Average bit widths are estimated for each approach, assuming 10K context length. Each column represents the average score for the tasks of each type.}
    \resizebox{40em}{!}{
    \begin{tabular}{c|l|c|c|c|c|c|c|c|c|c}
    \toprule
        \multirow{2}{*}{Model} & \multirow{2}{*}{Method} & Avg. & Comp. & Multi- & Single- & Summa- & \multirow{2}{*}{Few-Shot} & \multirow{2}{*}{Code} & \multirow{2}{*}{Synthetic} & \multirow{2}{*}{Avg.} \\
        & & Bits & Ratio & QA & QA & rization & & & & \\
    \midrule
        \multirow{6}{*}{Mistral-7B-v0.2} \ 
        & Baseline  & 16 & 1.00x & 29.63 & 36.43 & 28.10 & 66.71 & 54.16 & 44.87 & 42.54 \\
    \cmidrule{2-11}
        & Palu-30\% & 16 & 1.43x & 29.83 & 36.52 & 27.48 & 65.70 & 55.16 & 37.92 & 41.55 \\
        & Palu-50\% & 16 & 2.00x & 26.92 & 35.33 & 26.01 & 64.04 & 44.54 & 16.88 & 36.23 \\
        & KIVI-2 & 3.16 & 5.05x & 28.81 & 35.07 & 27.60 & 66.45 & 54.47 & 40.28 & 41.45 \\
        & Palu-30\% (3 bits) & 3 & 7.59x & 29.48 & 36.40 & 27.20 & 65.73 & 53.19 & 34.74 & 40.77 \\
        & Palu-50\% (3 bits) & 3 & 10.6x & 26.73 & 32.72 & 25.73 & 63.25 & 44.43 & 18.57 & 35.71 \\
    \midrule
        \multirow{6}{*}{LongChat-7B-v1.5} \ 
        & Baseline  & 16 & 1.00x & 23.95 & 31.12 & 26.74 & 63.80 & 56.91 & 15.25 & 36.32 \\
    \cmidrule{2-11} 
        & Palu-30\% & 16 & 1.43x & 22.42 & 29.43 & 25.52 & 62.87 & 58.99 & 14.25 & 35.45 \\
        & Palu-50\% & 16 & 2.00x & 22.61 & 25.33 & 22.73 & 60.12 & 43.52 & 6.84 & 30.82 \\
        & KIVI-2    & 3.16 & 5.06x & 23.24 & 30.19 & 26.47 & 63.54 & 53.51 & 16.13 & 35.60 \\
        & Palu-30\% (3 bits) & 3 & 7.59x & 23.12 & 29.21 & 25.04 & 61.99 & 54.38 & 11.25 & 34.33 \\
        & Palu-50\% (3 bits) & 3 & 10.6x & 18.56 & 24.14 & 22.35 & 58.76 & 40.50 & 6.02 & 29.03 \\
    \bottomrule
    \end{tabular}}
    \label{tab:longbench_appendix}
\end{table*}

\begin{figure}[h!]
\centering
    \includegraphics[width=1.0\columnwidth]{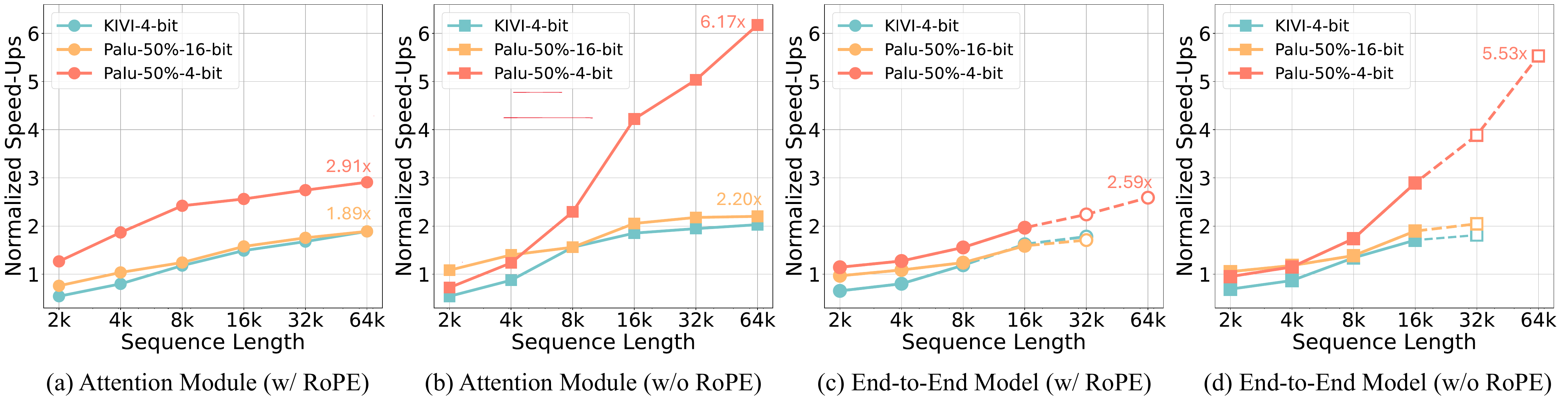}
    \caption{
    Normalized speedup for both the attention module and end-to-end model decoding. Solid lines represent exact measurements, while dashed lines indicate the FP16 baselines are out of memory, and the speedpus are compared the estimated baseline's latency.}
\label{fig:kernel_and_attention_latency}
\end{figure}

\vspace{-0.2cm}
\subsubsection{Speedups of Attention Module and End-to-end Decoding.}
\label{Exp: Speedups}

\paragraph{Attention module speedup.}
We compare latency against standard attention without compression or quantization and show the speedups of \palu{} and KIVI-4-bit in Fig. \ref{fig:kernel_and_attention_latency} (a) and (b) for RoPE-based and non-RoPE attention.
For RoPE-based attention, we applied our online reconstruction kernel for key and employed offline fusion for value as described in Sec \ref{sec:rope}.

As shown in Fig. \ref{fig:kernel_and_attention_latency} (a), 
\palu{} has minimal to no speedup when the sequence length is short, e.g. 4K.
However, as sequence length increases, \palu{} delivers substantial performance gains.
At 64K input length, \palu{} achieves a \textbf{1.89× speedup} over the FP16 baseline when using low-rank projection alone. 

By further applying 4-bit quantization to the Value states, the \textbf{speedup rises to 2.91$\times$} for the same 64K context length, owing to our optimized low-precision kernel and reduced memory loading times.
This performance notably surpasses KIVI-4-bit, which only achieves a 1.89× speedup at 64K, hindered by the overheads of its fined-grained group quantization.
Notably, for RoPE-based attention, \palu{}-4-bit does not quantize key, as our online reconstruction kernel only supports FP16 precision for now.

For non-RoPE attention, we apply matrix fusion to both the Key and Value states (Eq. \ref{eq:fused qk}), effectively eliminating all reconstruction overhead. 
At a 64K sequence length with a 50\% compression rate, \palu{} achieves a \textbf{2.20× speedup over the FP16 baseline}. 
By further applying 4-bit quantization to both the Key and Value states, \palu{} boosts the speedup to \textbf{6.17×} for 64K input length. 
These results demonstrate that combining low-rank compression and quantization significantly enhances inference efficiency, particularly in long-context scenarios.

\paragraph{End-to-end speedup.}
We present the end-to-end speedups in Fig. \ref{fig:kernel_and_attention_latency} (c) and (d), measuring the latency of generating the next token at various input lengths and comparing the results to the FP16 baseline. 
Similar to the attention performance results, \palu{} shows minimal or no speedup for short sequences but delivers significant acceleration for longer sequences. 
Without quantization, \palu{} achieves up to \textbf{1.71× and 2.05× speedups} for RoPE-based and non-RoPE models, respectively.
With a 50\% compression rate, \palu{} runs up to 32K input length on an RTX 4090 GPU. 
By incorporating 4-bit quantization, \palu{} handles even longer sequences and delivers \textbf{2.59× and 5.53× end-to-end speedups} at a 64K sequence length. 
\palu{} integrated with quantization provides a substantial speed advantage over KIVI-4-bit, which only reaches 1.78× and 1.81× speedups at 32K sequence length for RoPE and non-RoPE scenarios, respectively, and is out-of-memory for longer sequences.

\subsubsection{Kernel for RoPE-Based Attention Score} 
\label{sec: kernel speedup}

In this section, we evaluate the performance of our online reconstruction kernel for RoPE-based attention scores. We measure latency from the \textbf{pre-RoPE query vector} to \textbf{post-GEMV attention score}, and compare it with PyTorch's GEMV, which is used in the baseline attention (see Fig. \ref{fig: framework}).

\begin{wrapfigure}{r}{0.35\textwidth}
\vspace{-15pt}
\begin{center}
 \includegraphics[width=0.28\textwidth]{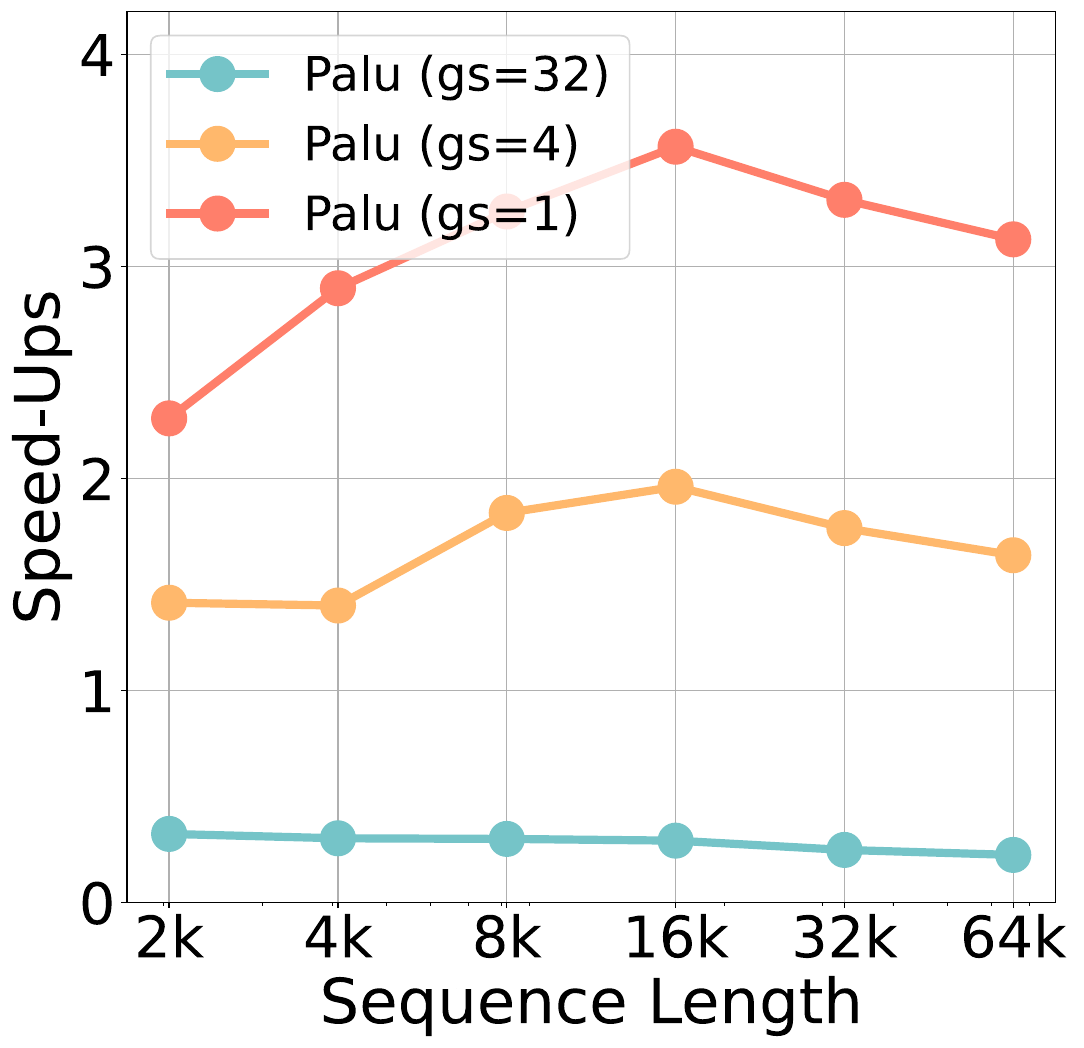}
 \caption{
    Speedup of \palu{}'s attention score kernel with online reconstruction.
    }\label{fig:kernel_speed_up}
\end{center}
\vspace{-15pt}
\end{wrapfigure}

We present speedups for group size 1, 4, and 32 at different sequence lengths in Fig. \ref{fig:kernel_speed_up}. For gs-32 (J-LRD), the highest accuracy decomposition, the high reconstruction cost causes a significant slowdown across all sequence lengths. For gs-1 (M-LRD), our kernel achieves up to a 3.56$\times$ speedup at sequence length 16K, showing strong performance when moderate accuracy loss is acceptable. For gs-4 (G-LRD), our kernel reaches up to 1.95× speedup. These results emphasize the need to explore various decomposition granularities for better accuracy and speed tradeoffs.

We also observe that speedup decreases for sequence lengths beyond 16K due to rising reconstruction costs, shifting the online reconstruction from memory- to compute-bound. A potential optimization is to quantize the decomposed weight matrices further and leverage high-throughput, low-precision hardware (\textit{e.g.,} INT4 Tensor Cores) for online reconstruction, which we leave for future work. Despite the speedup drop at longer lengths, \palu{}'s overall attention speedup increases with longer input, thanks to matrix fusion on the Value state and the reduced memory footprint.

\vspace{-5pt}
\section{Related Work}
\label{sec:related work}



\vspace{-3pt}
\paragraph{SVD for LLM Compression.}
Several works have explored using SVD to compress LLMs. An early approach \citep{NaiveSVD} applied standard SVD to weight matrices, resulting in significant compression errors. FWSVD \citep{fwsvd} addressed this by using Fisher information to prioritize parameters, while ASVD \citep{asvd} considered activation outliers. SVD-LLM \citep{svdllm} further minimized compression loss for each singular value. Unlike these methods, which compress model weights, \palu{} focuses on reducing \kv{} size. 
A concurrent work, \citep{CompressKVHead}, also explores \kv{} compression using low-rank projection, deriving low-rank matrices from \kv{} with calibration data and grouping attention heads similarly to \palu{}’s G-LRD. 
However, it requires LoRA finetuning after decomposition. 
In contrast, \palu{} directly decomposes the weight matrix, preserving accuracy without finetuning, and introduces additional innovations, including rank search, quantization integration, and optimized GPU kernels.

\vspace{-3pt}
\paragraph{\kv{} Compression.}
Quantization is a widely used technique for compressing \kv{}. Atom \citep{atom} applies simple per-token quantization, while WKVQuant \citep{wkvquant} introduces a two-level scheme to enhance accuracy. KIVI \citep{kivi} uses per-channel and per-token quantization for Keys and Values, combined with fine-grained group quantization at group size 32. KVQuant \citep{kvquant} employs a similar setup but incorporates non-uniform quantization and sparse matrices to handle outliers. On top of these approaches, GEAR adds a low-rank matrix to compensate for quantization errors. In \palu{}, we leverage low-rank techniques to exploit hidden dimension redundancy and, with simple per-token quantization, achieve outstanding compression results.

\vspace{-3pt}
\paragraph{MLA.} 
The recently released DeepSeek-V2 model \citep{deepseekai2024deepseekv2} introduces the MLA mechanism, which reduces \kv{} size by down-projecting Key and Value to a low-rank space and reconstructing them to full rank at runtime. 
Although MLA may seem similar to \palu{} at a high level, particularly with J-LRD, our design and derivation processes are fundamentally different. 
Unlike MLA, a new attention mechanism requiring pre-training, \palu{} is specifically designed for post-training integration. 
\palu{} focuses on converting existing models with MHA or GQA to support low-rank compressed \kv{}, preserving high accuracy while enhancing inference efficiency.

\vspace{-3pt}
\section{Conclusion}
\vspace{-3pt}


We introduce \palu{}, a novel \kv{} compression framework that decomposes linear projection weight matrices and caches the compressed latent representations. 
We propose various optimizations, including group-head low-rank decomposition, automatic rank allocation algorithm, quantization compatibility enhancement, and customized kernels with operator fusion. With these optimizations, \palu{} can maintain accuracy while achieving significant memory reduction and high inference speedup. Experiments show that, with 50\% low-rank compression and 4-bit quantization, \palu{} accelerates RoPE-based attention module by up to 2.91$\times$ and delivers up to 2.2$\times$ end-to-end speedup for the same RoPE-based model, while preserving strong accuracy on various benchmarks.

\bibliography{iclr2025_conference}
\bibliographystyle{iclr2025_conference}

\appendix
\newpage
\section*{Appendix}

\section{Quantization Basics}
Quantization techniques use discrete low-bit values to approximate high-precision floating points. The general asymmetric uniform quantization function is defined as:
\begin{equation}
    \overline{\mathbf{X}} = \text{clamp}\Big(\nint*{\frac{\mathbf{X}}{s}}+z, 0, 2^{B}-1), 
\end{equation}
where $\overline{\mathbf{X}}$ denotes the approximated tensor with low-bit representations (i.e., 4-bit integers), $\mathbf{X}$ is the floating-point tensor, $s= \frac{\mathbf{X}_\text{max} - \mathbf{X}_\text{min}}{2^{B} - 1}$ is the scaling factor, and 
$z=\nint*{\frac{-\mathbf{X}_\text{min}}{s}}$ is a zero-point. The $\nint*{\cdot}$ is the rounding operation. 

\section{Kernel Implementation Details}
\label{appendix: kernel impl}
\begin{figure}[h!]
\centering
    \includegraphics[width=0.65\textwidth]{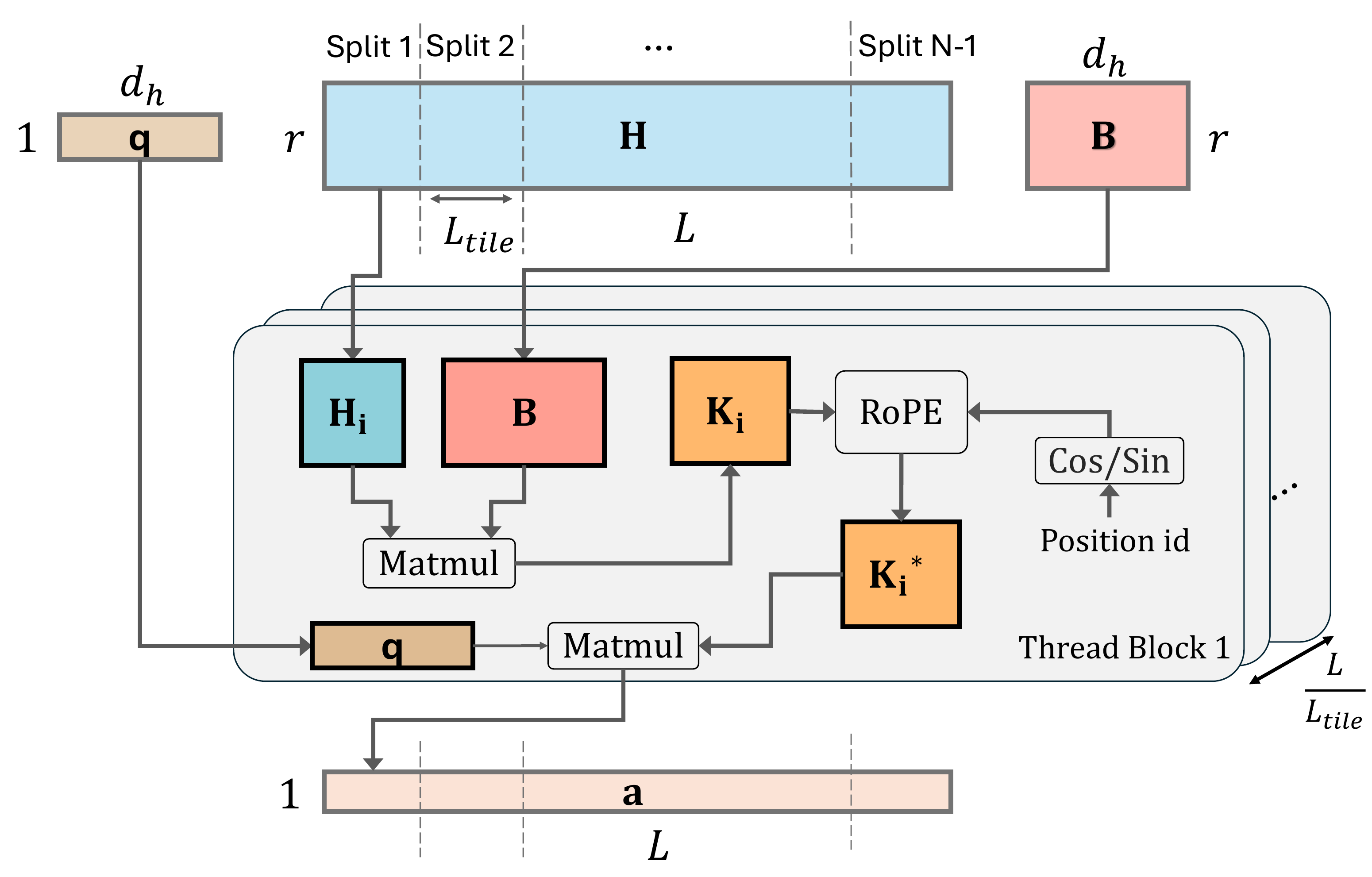}
    \caption{
    Illustration of our fused GPU kernel for computing attention scores with online reconstruction. In this figure, $\mathbf{q}$ represents the query vector, $\mathbf{H}$ denotes the low-rank compressed key states, and $\mathbf{B}$ stands for the reconstruction matrices.
    }
\label{fig:kernel_impl}
\end{figure}

\paragraph{Kernel for attention score calculation with reconstruction.} The central idea of \palu{} is to leverage low-rank latent representations to accelerate the attention mechanism by reducing data transfer overhead. Instead of working directly with the full-sized key matrix, we store and transfer a compressed low-rank latent representation, denoted as $\mathbf{H} \in \mathbb{R}^{L \times r}$. During computation, our custom GPU kernel performs an on-the-fly reconstruction using a reconstruction matrix $\mathbf{B} \in \mathbb{R}^{r \times d_h}$, producing a restored key matrix $\mathbf{K} \in \mathbb{R}^{L \times d_h}$, where $L$ is the sequence length, $d_h$ is the hidden dimension, and $r$ denote the remaining rank after performing low-rank projection. The query vector, represented as $\mathbf{q} \in \mathbb{R}^{1 \times d_h}$, then multiplies with the reconstructed keys to obtain the attention scores.

To efficiently leverage parallelism, we perform tiling along the sequence length dimension $L$. Specifically, we split the sequence into smaller tiles of size $L_{\text{tile}}$, assigning each tile to a dedicated thread block. Each thread block independently reconstructs a submatrix $\mathbf{H}_i \in \mathbb{R}^{L_{\text{tile}} \times d_h}$ from the low-rank latent representation $\mathbf{H}$, then applies the positional embedding using RoPE, and finally performs the matrix-vector multiplication between $\mathbf{q}$ and $\mathbf{H}_i$ to produce partial attention scores. This design ensures that all intermediate computations, from reconstruction to embedding and final multiplication, remain entirely in on-chip memory (\textit{i.e.,} share memory), thus minimizing high-latency memory access and taking full advantage of the GPU’s parallel processing capabilities to achieve significant speedups.

\section{Discussion regarding memory usage}
\label{append:lr_weight_size}
\paragraph{}{In this work, the experimental results focus on the compression rate of the \kv{} as a key metric. However, it is crucial to consider overall memory savings as a more significant factor. For instance, as demonstrated in Sec. \ref{subsec:SVD}, a typical compression rate of 30\% can lead to an increase in weight size by approximately 40\%. This increase is calculated under the assumption that $m = n$ and $r = 0.7n$, resulting in the equation \(\frac{mr+nr}{mn} = 1.4\). Such an increase indicates substantial extra memory usage. }

\paragraph{}{This issue primarily arises in J-LRD decomposition schemes, where the projections of all heads are decomposed jointly. In contrast, our M-LRD decomposition schemes and optimized G-LRD schemes involve non-square target matrices. For example, in the G-LRD scheme with a group size of 4, the target matrix is formed by concatenating the original projection matrices of each attention head in the group. In the Llama-2-7b model, with an embedding dimension of 4096 and head dimensions of 128, each projection matrix is 4096$\times$128, resulting in a concatenated matrix of size 4096$\times$512. In this case, the dimensions should be considered as $m = 8n$. Applying the referenced equation $\frac{mr + nr}{mn}$ with $r = 0.7n$, we find that $\frac{mr + nr}{mn} = 0.7875$, indicating no additional storage cost and, in fact, achieving an additional 21.25\% memory savings.}

\paragraph{}{Furthermore, it is important to highlight that the weights associated with the K and V projections account for only 2 out of 7 linear layers within transformer blocks, comprising merely 16\% of the parameters in Llama-2-7b models. This limits the overall impact on memory usage. Thus, while J-LRD may incur overhead, the M-LRD and G-LRD schemes provide efficient alternatives that do not lead to increased memory usage, making them viable options for practical applications.}

\section{Overall Memory Reduction}
\begin{figure}[h!]
\centering
    \caption{
    Total memory usage for various sequence lengths in Llama-2-7B. \lr{} compression is at 50\% for \palu{}.
    }
    \includegraphics[width=0.35\textwidth]{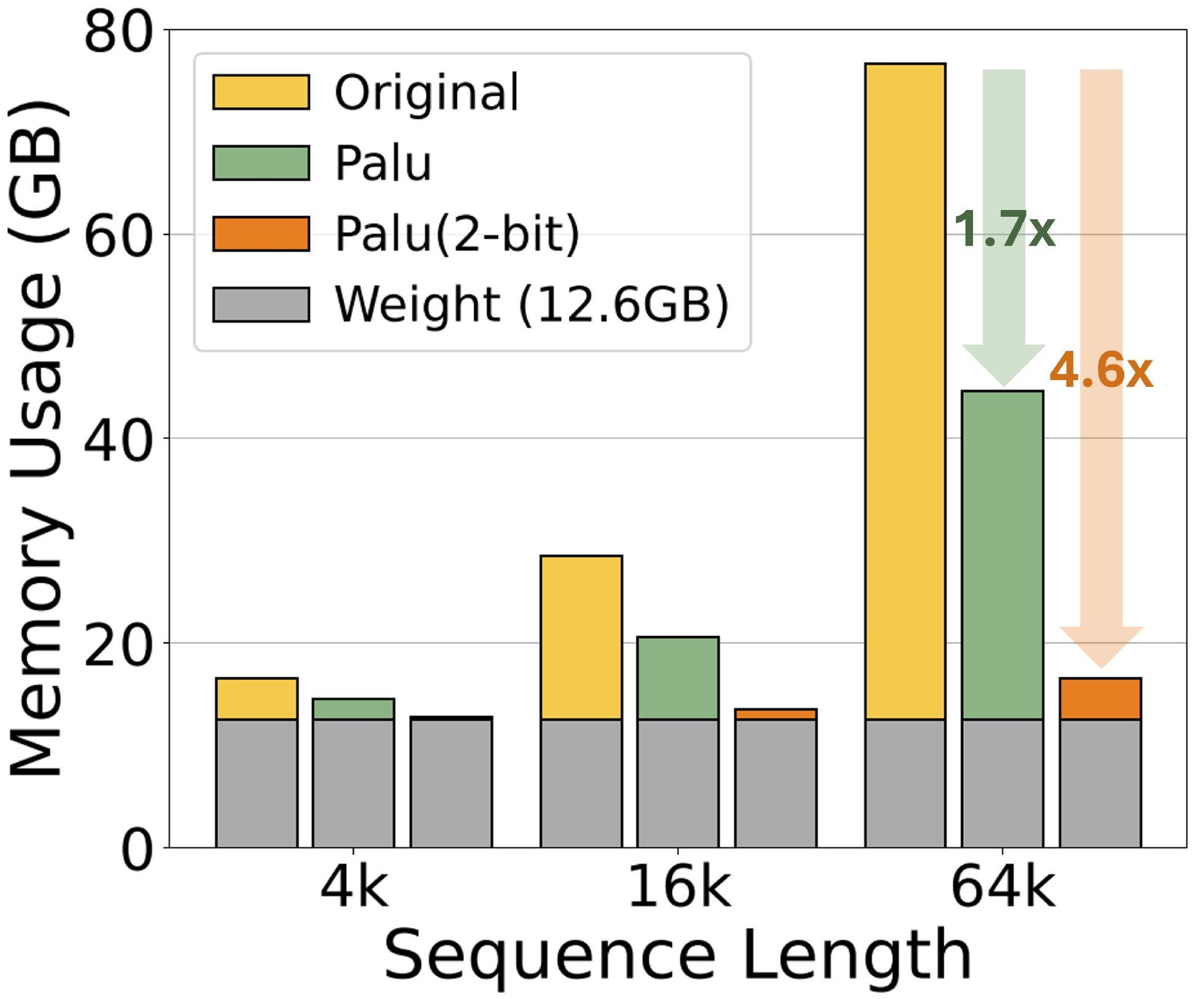}
\label{fig:memory}
\end{figure}

In Fig.~\ref{fig:memory}, we show the total memory usage, including model weights and \kv{} under different sequence lengths. We observe that \kv{} quickly becomes the memory bottleneck as sequence length grows.
At a sequence length of 64k, the \kv{} constitutes 78\% of the total memory consumption. 
With 50\% \lr{} compression, \palu{} effectively reduces memory usage by 1.7$\times$, and when combined with 2-bit quantization, it further decreases total memory usage by 4.6$\times$.

\begin{table*}[hbt]
\centering
\caption{Evaluation integrating LoRA with Palu on Llama-2-7B.} 
\resizebox{38em}{!}{
\begin{tabular}{ l | l | c  cccccccc }
\toprule
\multirow{2}{*}{Comp. Rate}      & \multicolumn{1}{c|}{\multirow{2}{*}{Method}}      & \multicolumn{8}{c}{Zero-Shot Accuracy (\%) $\uparrow$}     \\

& & OBQA & HellaSwag & PIQA & ARC-e & ARC-c & WinoGrande & Avg. & Diff. \\

\midrule

 Rate = 0\% & baseline        & 44.20 & 76.00 & 78.07 & 76.30 & 46.42 & 69.30 & 65.05 & - \\
\midrule
\multirow{3}{*}{\begin{tabular}{@{}l@{}}Rate = 50\%\\ w/o LoRA  \end{tabular}}
& J-LRD        & 45.40 & 75.57 & 77.48 & 75.97 & 45.31 & 69.22 & 64.83 & -0.22 \\
& G-LRD        & 43.60 & 73.39 & 76.33 & 73.02 & 42.57 & 66.77 & 62.61 & -2.44 \\
& M-LRD        & 39.60 & 65.35 & 74.76 & 67.17 & 35.24 & 64.64 & 57.79 & -7.26 \\
\midrule
\multirow{3}{*}{\begin{tabular}{@{}l@{}}Rate = 50\%\\ w/ LoRA  \end{tabular}}
& J-LRD         & 44.20 & 74.09 & 78.51 &	77.27 &	48.81 &	71.03 &	65.65 &	+0.60 \\
& G-LRD         & 43.40 & 73.08 & 78.56 &	75.72 &	47.10 &	69.85 &	64.62 &	-0.43\\
& M-LRD         & 41.80 & 70.78 & 78.02 &   73.86 & 43.86 & 69.22 & 62.92 & -2.12 \\

\bottomrule
\end{tabular}}
\label{tab:lora}
\end{table*}

\section{Integrating \palu{} with LoRA Finetune}

LoRA \citep{hu2022lora} has become one of the most widely used efficient fine-tuning techniques for adapting models to particular tasks or domains with limited data. It has also been applied with LLM compression approaches \citep{svdllm, LLMPruner} as a post-compression recovery technique to recover information loss after compression. In \palu{}, LoRA is also applicable to boost the accuracy further. 

To integrate LoRA with \palu{}, we introduce additional low-rank matrices $\mathbf{A}_r'\in \mathbb{R}^{d\times r'}$ and $\mathbf{B}_r'\in\mathbb{R}^{r'\times d}$ to refine the original low-rank projection as below:
\begin{equation}
    \mathbf{h} = \mathbf{A}\mathbf{x} + \mathbf{A}_{\mathbf{r'}}\mathbf{B}_{\mathbf{r'}}\mathbf{x}
\end{equation}
Here, $\mathbf{A}$ will be fixed parameters derived from low-rank decomposition from pre-trained weights of linear layers, while $\mathbf{A}_r'$ and $\mathbf{B}_r'$ are trainable parameters to capture the task-specific nuances and recovers the information lost during the compression. 

\paragraph{Setup.} Following \citealt{SliceGPT}, we sample 8k samples from the Alpaca training dataset as a fine-tuning dataset and apply LoRA with rank $r'=32$ and $\alpha=32$. All other hyper-parameters are aligned with \citet{SliceGPT}, except for the learning rate $2e-4$, and the use of a cosine learning rate scheduler.

\paragraph{Experiment Results.} We present the experiment results with LoRA in Table \ref{tab:lora}. Following \citealt{SliceGPT} With LoRA incorporated, J-LRD continues to show minimal performance degradation with an average drop of 1.00\%. G-LRD (gs=4) and M-LRD show improved results compared to their non-LoRA counterparts, with average drops of 2.01\% and 5.14\%, respectively. Notably, with LoRA integration, G-LRD shows only a 1.03\% accuracy difference compared to J-LRD.

\section{More Results on Zero-shot Accuracy}
\label{appendix:more_results}
Following up Sec. \ref{sec:exp_main_results}, we further report the perplexity and zero-shot evaluation results of \palu{} on the Llama-2-13B at 50\% compression rate. As shown 
in Table. \ref{tab:llama2-13b_zero_shot_and_ppl}, we observe that \palu{} achieve competitive accuracy drops around 3\% or less across different using either J-LRD, G-LRD, or M-LRD. Thus, the users may adopt M-LRD first to optimize the efficiency further. 
\begin{table*}[t!]
\centering
\caption{Perplexity and zero-shot accuracy of \palu{}, with different decomposition strategies at 50\%
}
\resizebox{37em}{!}{
\begin{tabular}{c | l | rr | ccccccc}
\toprule
\multirow{2}{*}{Model}      & \multicolumn{1}{c|}{\multirow{2}{*}{Method}} & \multicolumn{2}{r|}{Perplexity $\downarrow$}        & \multicolumn{7}{c}{Zero-Shot Accuracy (\%) $\uparrow$}     \\
                            & \multicolumn{1}{c|}{}                        & \multicolumn{1}{r}{Wiki2} & \multicolumn{1}{r|}{C4} & OBQA  & Hella & PIQA  & ARC-e & ARC-c & Wino  & Avg.  \\ \midrule
\multirow{4}{*}{Llama-2-13B} & Baseline                                     & 4.88                      & 6.70                    & 45.20 & 79.39 & 79.11 & 79.42 & 49.06 & 72.38 & 67.43 \\ \cmidrule{2-11} 
                            & J-LRD                                        & 4.97                      & 6.92                    & 46.40 & 79.48 & 78.62 & 79.29 & 49.91 & 70.56 & 67.38 \\
                            & G-LRD                                        & 5.31                      & 7.76                    & 45.60 & 77.29 & 77.42 & 76.05 & 45.99 & 72.45 & 65.80 \\
                            & M-LRD                                        & 5.65                      & 8.34                    & 43.20 & 74.34 & 77.53 & 75.76 & 45.39 & 68.98 & 64.20 \\ \bottomrule
\end{tabular}}
\label{tab:llama2-13b_zero_shot_and_ppl}
\end{table*}

\begin{table}[h!]
\caption{
Ablation study of low-rank decomposition group size on perplexity for the Llama-2-7B model at a 50\% compression rate using Wikitext-2.
}
\centering
\resizebox{13em}{!}{
%
%
\begin{tabular}{ccc}
\toprule
Method                 & \multicolumn{1}{l}{Group Size} & \multicolumn{1}{l}{Perplexity} \\
\midrule
Baseline               & \multicolumn{1}{c}{-}          & 5.47                           \\ 
\midrule
J-LRD                  & 32                             & 5.62                           \\ 
\midrule
\multirow{4}{*}{G-LRD} & 16                             & 5.74                           \\
                       & 8                              & 5.88                           \\
                       & 4                              & 6.01                           \\
                       & 2                              & 6.42                           \\ \midrule
M-LRD                  & 1                              & 6.81                           \\
\bottomrule
\end{tabular}}
\label{tab:group_size}
\end{table}

\section{Ablation Study}
\subsection{Influence of Different Group Size}
Since our proposed G-LRD method allows for balancing performance and efficiency by adjusting the group size, we conducted an ablation study on group size. As seen in Table \ref{tab:group_size}, as the group size increases, the amount of shared information also increases, leading to improved performance.

\subsection{Influence of Walsh-Hadamard Transform}
We conduct the ablation study to profile the benefits of applying the Walsh-Hadamard Transform (WHT). Experiment results are reported at Table \ref{tab:quantization_ablation}. On the 3-bit quantization level, we observe that the Hadamard Transform only brings a slight amount of perplexity. However, when we quantize the low-rank representation more extremely (\textit{i.e.,} 2-bit), we can observe a notable 4.17 perplexity enhancements. It's worth re-emphasizing that Hadamard Transform will not bring extra overhead during inference, as \palu{} optimizes the WHT process via offline preprocessing. The reader may refer to Sec. \ref{sec:Low-Rank Aware Quantization} for more details. 

\subsection{Automatic Rank Allocation vs. Uniform Rank Allocation}
\label{append:detailed_key_value_rank}
\begin{figure}[h]
\centering
    \includegraphics[width=0.47\textwidth]{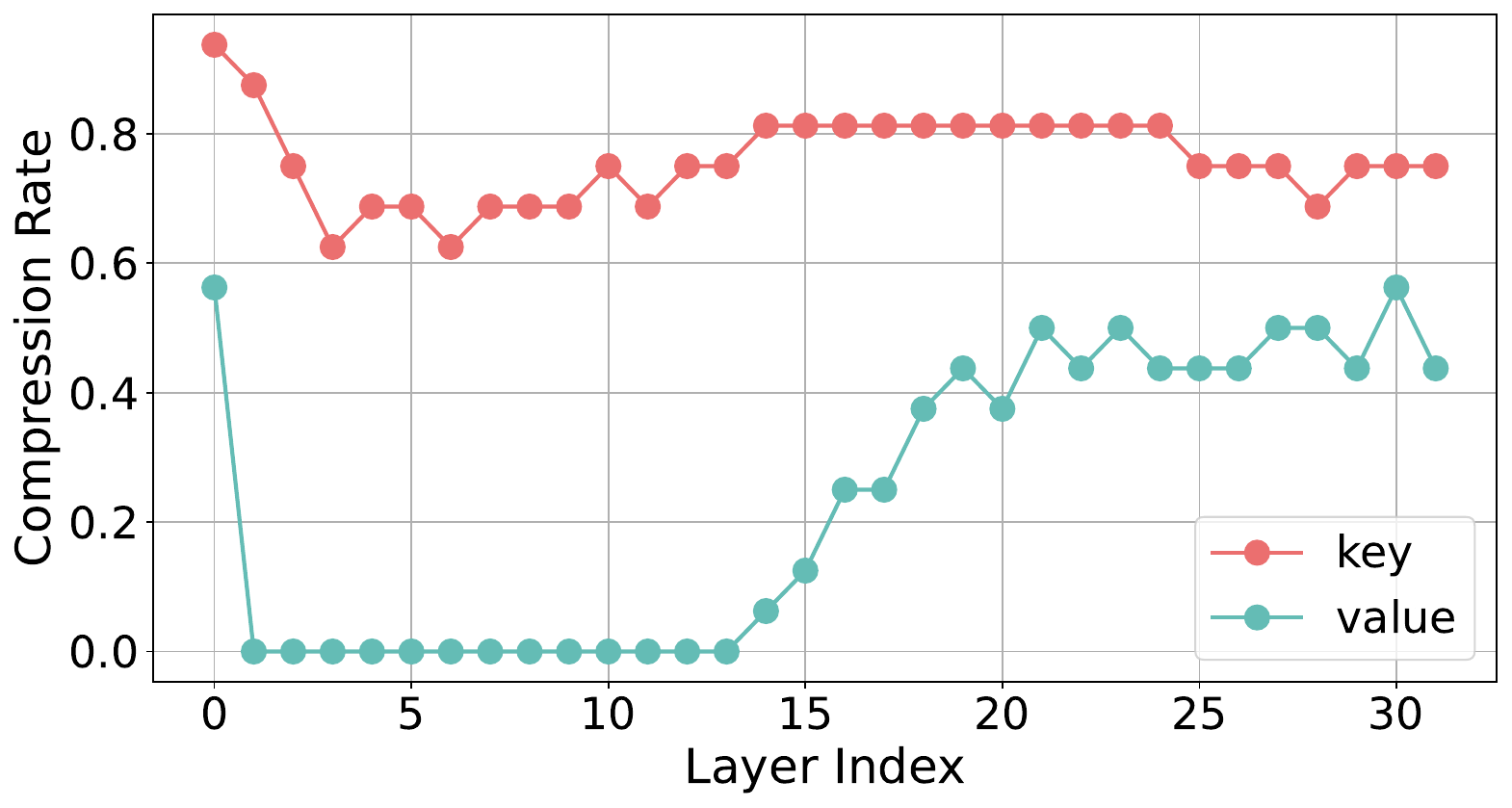}
    \caption{
        Visualization of layer-wise low-rank compression rate on Llama-2-7B with 50\% of overall compression rate. Here, compression rates (\textit{i.e.,} rank) are allocated using the proposed Fisher Information-based automated rank allocation algorithm. 
    }
    \label{fig: rank viz}
\end{figure}

Table \ref{tab:searching_ablation} presents the ablation study on the impact of different rank allocation schemes on the model's accuracy. Applying rank searching results in a notable performance improvement. For instance, at a compression rate of 50\%, there is a significant reduction in perplexity by 2.18. Fig. \ref{fig: rank viz} visualizes the rank allocation across different transformer blocks for key and value projection layers. The results clearly demonstrate a non-uniform allocation result. Specifically, we observe that the value is generally allocated a higher rank than the key. Additionally, the first half of the layers are assigned higher ranks, indicating their greater importance in preserving model performance. This visualization underscores the effectiveness of our rank search algorithm in identifying and allocating appropriate ranks to different components, thereby optimizing the balance between compression and accuracy.

\begin{table}[hbt]
\centering
\caption{Ablation Study on different quantization settings for quantizing low-rank latent representations. Same as Sec. \ref{sec:exp_quantization}, we use the WikiText-2 with sequence length set to 4096 as the evaluation benchmark.
}
\begin{tabular}{l@{\hskip 0.3in} l}
\toprule
Method & Wikitext-2 PPL $\downarrow$ \\
\midrule
Llama2-7B &  5.12 \\
\midrule
Palu-30\% (FP16) & 5.25 \\
\hspace{1em} + 3-bits w/o Hadamard & 5.52 \\
\hspace{1em} + 3-bits w Hadamard & 5.33 (0.19$\downarrow$) \\
\noalign{\vskip -6pt}
\multicolumn{2}{l}{\hspace{1em}\rule{0.5\linewidth}{0.1pt}} \\
\hspace{1em} + 2-bits w/o  Hadamard & 9.48 \\
\hspace{1em} + 2-bits w  Hadamard & 5.77  (3.71$\downarrow$)\\
\midrule[1.2pt]
Palu-50\% (FP16) & 5.63 \\
\hspace{1em} + 3-bits w/o Hadamard & 5.99 \\
\hspace{1em} + 3-bits w Hadamard  & 5.77 (0.22$\downarrow$) \\
\noalign{\vskip -6pt}
\multicolumn{2}{l}{\hspace{1em}\rule{0.5\linewidth}{0.1pt}} \\
\hspace{1em} + 2-bits w/o  Hadamard & 10.58 \\
\hspace{1em} + 2-bits w  Hadamard & 6.41 (4.17$\downarrow$) \\
\bottomrule[1.2pt]
\end{tabular}
\label{tab:quantization_ablation}
\end{table}

\begin{table}[H]
\centering
\caption{Ablation study on w/ and w/o rank search. We use Llama-2-7b and Wikitext-2 with sequence length 2048 as the benchmark.
}
\resizebox{24em}{!}{
\begin{tabular}{l@{\hskip 0.2in} l l l }
\toprule
    & Rate=30\%  & Rate=50\% & Rate=70\% \\
\midrule
Uniform &   6.34   &   7.36   & 10.77 \\
Automatic (ours)  &    5.62 (0.72$\downarrow$)  &   6.02  (1.36$\downarrow$) & 8.59 (2.18$\downarrow$) \\
\bottomrule
\end{tabular}}
\label{tab:searching_ablation}
\end{table}

\section{Experiment Details}
\label{append:exp_setting_details}
\subsection{Zero-Shot Evaluation Details}
We selected six zero-shot tasks from the LM-eval benchmark to evaluate \palu{}:
\begin{itemize}[leftmargin=2em, itemsep=0.3em, parsep=0.1em, topsep=0.2em]
    \item OpenBookQA (accuracy, \citeauthor{openbookqa})
    \item HellaSwag (acc\_norm, \citeauthor{hellaswag})
    \item PIQA (accuracy, \citeauthor{piqa})
    \item ARC-Easy (accuracy, \citeauthor{arc})
    \item ARC-Challenge (acc\_norm, \citeauthor{arc})
    \item WinoGrande (accuracy, \citeauthor{winogrande})
\end{itemize}
We report accuracy for WinoGrande, PIQA, and ARC-Easy, and accuracy normalized by sequence length (acc\_norm) for HellaSwag and ARC-Challenge.

\subsection{LongBench Evaluation Details}
For the LongBench evaluation in the manuscript, we selected eight tasks from four subgroups, ensuring a comprehensive evaluation of \palu{}. The tasks and their corresponding metrics are detailed below:

\begin{itemize}[leftmargin=1.4em, itemsep=0.3em, parsep=0.1em, topsep=0.2em]
    \item Single-Document QA:
    \begin{itemize}[leftmargin=1.4em, itemsep=0.3em, parsep=0.1em, topsep=0.2em]
        \item Qasper (F1 score, \citeauthor{Qasper})
    \end{itemize}

    \item Summarization:
    \begin{itemize}[leftmargin=1.4em, itemsep=0.3em, parsep=0.1em, topsep=0.2em]
        \item QMSum (ROUGE score, \citeauthor{QMSum})
        \item MultiNews (ROUGE score, \citeauthor{MultiNews})
    \end{itemize}

    \item Few-shot Learning:
    \begin{itemize}[leftmargin=1.4em, itemsep=0.3em, parsep=0.1em, topsep=0.2em]
        \item TREC (classification score, \citeauthor{TREC})
        \item TriviaQA (F1 score, \citeauthor{TriviaQA})
        \item SAMSum (ROUGE score, \citeauthor{SAMSum})
    \end{itemize}

    \item Code Completion:
    \begin{itemize}[leftmargin=1.4em, itemsep=0.3em, parsep=0.1em, topsep=0.2em]
        \item LCC (similarity score, \citeauthor{LCC})
        \item RepoBench-P (similarity score, \citeauthor{RepoBench})
    \end{itemize}
\end{itemize}
During the evaluation, we set the maximum sequence length to 31500 for both the Mistral and LongChat model.

\end{document}